\documentclass{article}

\PassOptionsToPackage{table,dvipsnames,xcdraw}{xcolor}

\usepackage{amsmath,amsfonts,bm}

\def\Figref#1{Figure~\ref{#1}}

\def\Secref#1{Section~\ref{#1}}

\def\eqref#1{equation~\ref{#1}}

\def\1{\bm{1}}

\DeclareMathAlphabet{\mathsfit}{\encodingdefault}{\sfdefault}{m}{sl}
\SetMathAlphabet{\mathsfit}{bold}{\encodingdefault}{\sfdefault}{bx}{n}

\usepackage[english]{babel}
\usepackage{blindtext}
\usepackage{xspace}
\usepackage{amsmath}
\usepackage[normalem]{ulem}
\usepackage{caption}
\usepackage{graphicx}
\usepackage{booktabs}
\usepackage{arydshln} %
\usepackage{xurl}
\usepackage{wrapfig}
\usepackage[toc,title]{appendix}
\usepackage{afterpage}
\usepackage{tikz}
\usepackage{enumitem}
\usepackage{stfloats}

\usepackage{comment}
\usepackage{algorithm}
\usepackage[noend]{algpseudocode}
\usepackage{multirow}

\usepackage{xspace}
\usepackage{mfirstuc} %
\usepackage{longtable}
\usepackage{listings}

\usepackage[most]{tcolorbox} %
\usepackage{graphicx} %
\usepackage{multicol} 
\usepackage{subcaption}

\usepackage[affil-it]{authblk} 
\usepackage{amsmath,amsfonts,bm}

\newcommand{\sys}{\textit{Cake}\xspace}
\newcommand{\kv}{\text{KV cache}\xspace}

  {\begin{list}{\labelitemi}{\itemsep2pt \topsep2pt \parsep0.00in
  \partopsep=0pt \leftmargin1.2em}}%
  {\end{list}}

\let\latexusecounter=\usecounter
\def\compactsortof{\itemsep=2pt \topsep=2pt \parsep=0.00in \partopsep=0pt
\leftmargin=1.2em}

\newcommand{\BULLET}{\vspace{+.00in} \noindent $\bullet$ \hspace{+.00in}}

\newcommand{\mysection}[1]{\vspace{-.06in}\section{#1}\vspace{-.02in}}
\newcommand{\mysubsection}[1]{\vspace{-.06in}\subsection{#1}\vspace{-.03in}}

\usepackage{enumitem}
\setlist[itemize]{nosep, itemsep=0pt, topsep=0pt, leftmargin=1em, labelwidth=*, align=left}
\setlist[enumerate]{nosep, itemsep=0pt, topsep=0pt, leftmargin=1em, labelwidth=*, align=left}

\usepackage[accepted]{icml2025}

\usepackage{amsmath}
\usepackage{amssymb}
\usepackage{mathtools}
\usepackage{amsthm}
\usepackage{adjustbox}

\usepackage[capitalize,noabbrev]{cleveref}

\theoremstyle{plain}

\theoremstyle{definition}

\theoremstyle{remark}

\begin{document}

\twocolumn[
\icmltitle{Compute Or Load KV Cache? Why not Both?}

\vspace{-0.15in}

\icmlsetsymbol{equal}{*}

\begin{icmlauthorlist}
\icmlauthor{Shuowei Jin}{equal}
\icmlauthor{Xueshen Liu}{equal}
\icmlauthor{Qingzhao Zhang}{}
\icmlauthor{Z. Morley Mao}{}
\end{icmlauthorlist}

\centerline{University of Michigan}

\icmlcorrespondingauthor{Shuowei Jin}{jinsw@umich.edu}
\icmlcorrespondingauthor{Xueshen Liu}{liuxs@umich.edu}

\icmlkeywords{Efficient LLM Inference, KV Cache}

\vskip 0.2in
]

\printAffiliationsAndNotice{\icmlEqualContribution}  %

\begin{abstract}

Large Language Models (LLMs) are increasingly deployed in large-scale online services, enabling sophisticated applications. However, the computational overhead of generating key-value (KV) caches in the prefill stage presents a major bottleneck, particularly for long-context inputs. Prefix caching mitigates this issue by storing KV caches for reuse, reducing redundant computation. Despite its advantages, prefix caching suffers from high latency due to the limited I/O bandwidth of storage devices, constraining inference efficiency. To address this challenge, we introduce \textbf{\sys}, a novel KV cache loading system that optimally utilizes both computational and I/O resources in parallel. \sys employs a bidirectional scheduling strategy that dynamically balances KV cache computation and loading, ensuring efficient resource utilization. Additionally, \sys incorporates an adaptive scheduling mechanism that seamlessly integrates with non-prefix caching requests, improving system throughput and adapting to fluctuating resource availability.  Through extensive evaluations across various hardware configurations, datasets, and storage conditions, \sys achieves on average \textbf{2.6×} reduction in Time to First Token (TTFT) compared to compute-only and I/O-only methods. Our findings highlight \sys as an effective and practical solution for optimizing long-context LLM inference, bridging the gap between computation and I/O efficiency in large-scale AI deployments.  
\end{abstract}

\mysection{Introduction}

Large Language Models (LLMs) have become a cornerstone of modern AI applications, powering a wide range of large-scale online services. As these models are increasingly adopted, ensuring efficient online inference has emerged as a critical research and engineering challenge~\cite{kwon2023efficient, agrawal2024taming, zheng2023efficiently, miao2024specinfer, leviathan2023fast, ning2023skeleton, jin2024adaptive}. Recent advancements in LLMs, such as the expansion of context windows~\cite{gpt4o, claude}, have enabled sophisticated applications including long document understanding~\cite{wang2024leave}, long-context Retrieval-Augmented Generation (RAG)~\cite{jiang2024longrag}, and the creation of complex LLM-based agents~\cite{zhang2024chain}. However, while these capabilities enhance utility, they introduce substantial computational overhead, particularly during the inference stage, due to the cost of generating key-value (KV) caches.~\footnote{\kv is widely used in state-of-the-art inference systems to reduce computational overhead during decoding for each request.}
For example, generating the \kv for a 72K-token input (e.g., a 200-page book such as The Great Gatsby) using Llama2-70B on an NVIDIA A100 GPU takes approximately 30 seconds, significantly impacting user experience. Given this computational bottleneck, strategies to optimize LLM inference workflows are essential.

In real-world applications, many tokens are reused across users and conversations, presenting an opportunity for system optimization. For instance, in multi-turn conversations, follow-up queries reuse the key-value (\kv) pairs from prior tokens, while in Retrieval-Augmented Generation (RAG) workflows, document \kv can be shared across multiple user queries~\cite{jin2024ragcache, chan2024don}. To reduce these redundancies, \textbf{prefix caching} stores previously computed \kv and loads them into GPU memory for inference when the corresponding tokens are hit. Leading LLM service providers, including OpenAI, Anthropic, and DeepSeek, have integrated prefix caching mechanisms into their inference systems, lowering inference costs by over 50\%~\cite{openai2024promptcaching, anthropic2023promptcaching, deepseek2024promptcaching}.

\begin{figure}
    \centering
    \includegraphics[width=0.8\linewidth]{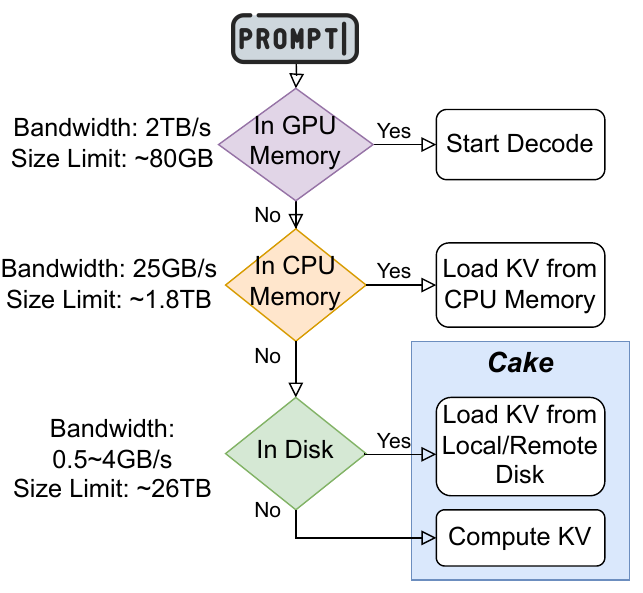}
    \caption{Workflow of long-context LLM inference with prefix caching. \sys operates during the \kv loading phase (highlighted in blue). The configuration parameters are based on the specifications of a LambdaLab GPU server~\cite{lambda}.}
    \label{fig:cake-scenario}
\end{figure}

Despite its advantages, deploying prefix caching at scale poses a key challenge: designing a high-capacity \kv management system across heterogeneous memory layers while ensuring low loading latency for optimal \textbf{Time-to-First-Token (TTFT)} during inference.
State-of-the-art inference engines~\cite{kwon2023efficient, zheng2023efficiently} typically store \kv in GPU and CPU memory, ensuring minimal latency. However, GPU and CPU memory are both limited and expensive, often requiring the eviction of \kv for long-context requests, making them impractical for large-scale inference services. To address this limitation, hierarchical storage systems have been proposed, leveraging CPU memory, local disks, and remote storage for managing \kv caches~\cite{gao2024cost, liu2023cachegen, yao2024cacheblend}. Figure~\ref{fig:cake-scenario} illustrates the hierarchical workflow of LLM inference with prefix caching. Upon receiving a query, the system searches for reusable \kv caches across three storage tiers:

\begin{itemize}
    \item \textbf{GPU Memory}: The fastest but most capacity-limited option (bandwidth: $\sim$2TB/s, size: $\sim$80GB). If the \kv cache is found, decoding proceeds immediately.
    \item \textbf{CPU Memory}: Offers greater capacity (bandwidth: $\sim$25GB/s, size: $\sim$1.8TB) but is slower than GPU memory.
    \item \textbf{Disk Storage}: The largest capacity option (bandwidth: 0.5--4GB/s, size: $\sim$26TB) but also the lowest bandwidth.
    \item \textbf{Compute}: If no \kv cache is found across all levels, the system computes it from scratch.
\end{itemize}

Due to the substantial capacity differences in the storage hierarchy, a significant portion of \kv caches reside on local or remote disks. As evaluated in AttentionStore~\cite{gao2024cost}, approximately 80\% of cache hits occur at the disk level. However, fetching large \kv from disk is constrained by the low I/O bandwidth of PCIe or network-based remote storage, significantly impacting TTFT, which adversely affects user experience.

\textbf{Contribution.} In this paper, we introduce \sys (\underline{C}omputation and I/O \underline{A}ware \underline{\kv} Cach\underline{E} loader), a novel \kv loading system designed to minimize latency when loading \kv from high-capacity, low-bandwidth storage layers. \sys optimally leverages the distinct characteristics of both computational and I/O resources in parallel, significantly reducing TTFT by 2.6x on average in long-context prefix caching scenarios.  

In addition to reducing latency, \sys is designed considering real-world practicality. We propose an adaptive scheduling mechanism that efficiently shares compute resources with non-prefix caching requests, improving overall throughput by 26\%. Additionally, \sys dynamically adjusts to fluctuations in network and computational resources, ensuring consistently optimal latency.

Furthermore, we conduct extensive experiments across diverse setups and provide a detailed analysis of \sys's effectiveness. Our findings offer practical insights into when \sys achieves the greatest performance gains, serving as a valuable guide for future real-world deployments.

To the best of our knowledge, \sys is the first system to demonstrate that efficiently utilizing both computational and I/O resources can optimally reduce TTFT in long-context prefix caching scenarios. Prior approaches either rely exclusively on computation or solely on I/O for \kv loading~\cite{liu2023cachegen,kwon2023efficient}, leaving a significant gap in understanding how to leverage the unique characteristics of both computation and I/O for efficient \kv cache loading. \sys addresses this gap, providing a comprehensive and adaptive solution for long-context prefix caching scenarios.

\mysection{Related Work}
\label{sec:related-work}

Previous work has explored two primary directions to develop practical prefix caching systems: (1) algorithm-level compression techniques that reduce the size of \kv, thereby decreasing loading time, and (2) system-level \kv management strategies that expand cache capacity across heterogeneous memory layers.

\textbf{\kv Compression.} Most work compress the \kv through quantization, token pruning, and model architectural modifications. Quantization methods~\cite{hooper2024kvquant,kang2024gear,liu2024kivi} reduce the precision of \kv representations while maintaining accuracy. Token pruning approaches like LLMLingua~\cite{jiang2023llmlingua}, ScissorHands~\cite{liu2024scissorhands}, and H2O~\cite{zhang2023h2o} identify and remove less important tokens from the \kv. At the model architecture level, Grouped-Query Attention (GQA)~\cite{ainslie2023gqa} reduce memory footprint by sharing key-value heads across queries in the group, while Multi-head Latent Attention (MLA) compresses \kv into compact latent vectors to reduce \kv size. 

\textbf{System Optimizations.} vLLM~\cite{kwon2023efficient} and SGLang~\cite{zheng2023efficiently} mainly optimize \kv management between GPU DRAM and CPU memory, enabling low-latency loading despite limited capacity. CachedAttention~\cite{gao2024cost} extends this by employing a hierarchical \kv management system across memory and disk mediums, effectively support multi-turn conversations. CacheGen~\cite{liu2023cachegen} addresses scenarios where \kv is stored in remote data storages, applying adaptive compression methods to reduce the latency of loading through network.

While these approaches focus on reducing the I/O load latency to reduce TTFT, our work tackles the problem from a different angle: combining computation and I/O in parallel to further reduce latency. It is an orthogonal design to existing methods.

\mysection{Background}
\label{sec:background}

\textbf{Attention and \kv.}  
The attention mechanism~\cite{vaswani2017attention} is a core component of LLMs, allowing them to model token relationships efficiently. \kv is a technique to improve attention module inference efficiency.

Given an input sequence \(X\), the attention module first transforms it into queries, keys, and values:  

\[
Q = XW_Q, \quad K = XW_K, \quad V = XW_V
\]

where \(W_Q\), \(W_K\), and \(W_V\) are learned projection matrices. The attention scores are then computed as:  

\[
\text{Attention}(Q, K, V) = \text{softmax}\left(\frac{QK^T}{\sqrt{d_k}}\right)V
\]

where \(d_k\) is the key dimension.  

During autoregressive decoding, this computation repeats at each step \(t\), generating a new token \(x_t\) based on previous tokens \(x_{<t}\). However, the key and value vectors for \(x_{<t}\) remain unchanged across decoding steps. To eliminate redundant computation, modern systems cache these as \( \text{past\_K} \) and \( \text{past\_V} \), computing only the query for \(x_t\) along with the new key \(k_t\) and value \(v_t\). This optimization, known as \kv, significantly reduces computational overhead.

\textbf{Chunk Prefill.}  
Chunk prefill is a technique used to optimize LLM inference, particularly for long input sequences. State-of-the-art LLM inference engines~\cite{kwon2023efficient,zheng2023efficiently} split inference into two phases:  
\begin{itemize}
    \item \textbf{Prefill:} Computes the KV cache for the input prompt.  
    \item \textbf{Decode:} Generates tokens sequentially using the cached KV, updating it incrementally.  
\end{itemize}

For long-context inputs, the prefill stage is highly compute-intensive, requiring substantial GPU resources for a long period. Chunk prefill~\cite{agrawal2023sarathi,agrawal2024taming} mitigates this by dividing the input sequence into smaller chunks to prefill it chunk by chunk and batching small chunk prefill together with decode requests. To be specific, vLLM forms a batch of requests for each inference step based on a predetermined token budget. The scheduler prioritizes decode requests, allocating one token from the budget to each. Any remaining tokens in the budget are then assigned to prefill requests for chunk prefill. This prevents long prefill operations from blocking decode requests, improving GPU utilization by co-locating compute-bound (prefill) and memory-bound (decode) tasks in the same batch. Currently, it is the default suggested scheduling mode in vLLM~\cite{kwon2023efficient}. 

We design \sys upon this approach, scheduling chunks for either computation or I/O. Compared to token-level scheduling, chunk-level scheduling better exploits GPU parallelism for maximum efficiency. Compared to sequence-level scheduling, it is more fine-grained, allowing greater flexibility for optimal scheduling strategies.

\mysection{Design of \sys}
\label{sec:design}

\begin{figure*}[t]
    \centering
    \includegraphics[width=0.8\linewidth]{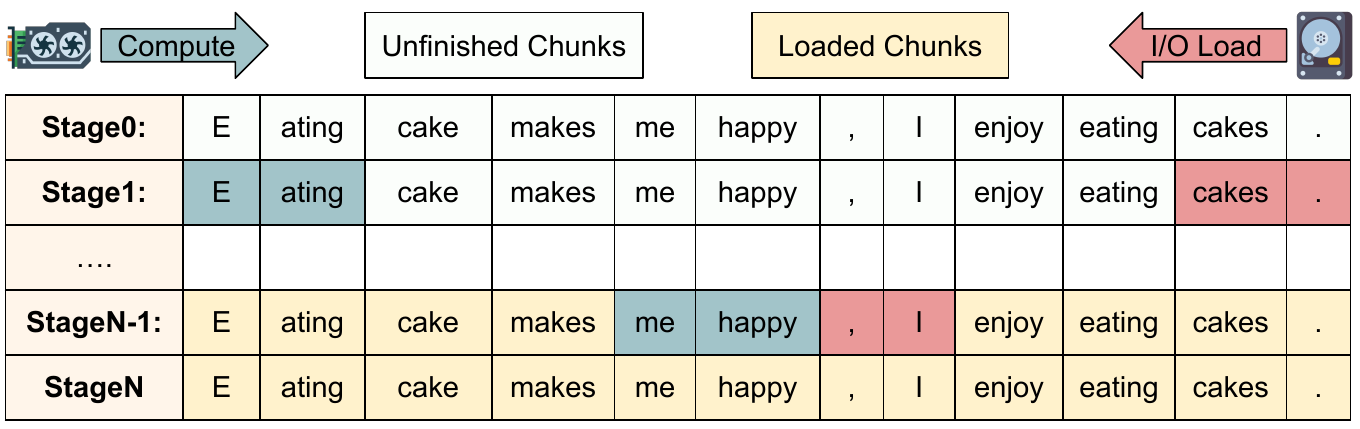}
    \caption{Workflow of \sys: Computation starts from the beginning of the sequence, while I/O loading starts from the end. Both processes progress in parallel and merge in the middle, ensuring efficient \kv loading and minimal latency.}
    \label{fig:cake-diagram}
\end{figure*}

We design \sys to optimize the latency of loading \kv from storage layers with limited bandwidth but large capacity, such as local disks and remote storage —where most cache hits occur in prefix caching scenarios. \sys leverages both compute and I/O resources bidirectionally and in parallel to minimize this latency. We present \sys workflow in \Figref{fig:cake-diagram}.

In this section, we first analyze the compute and I/O capability in common inference server setups. Next, we present key insights into \sys’s scheduling design, focusing on how to efficiently utilize the unique characteristics of compute and I/O. Finally, we discuss in real-world scenarios how \sys dynamically adapts to fluctuations in compute and I/O bandwidth, ensuring optimal latency.

\textbf{Part 1: Analysis of Compute and I/O Capability.} We first analyze the performance characteristics of \kv loading and computation across common inference system configurations. We evaluate the equivalent throughput (calculated as \kv file size divided by processing time) using vLLM~\cite{kwon2023efficient} with a chunk size of 512 tokens on the LongAlpaca-7B model~\cite{chen2023longlora}. Our experiments use various GPUs to process a random context of 32k tokens. A more detailed analysis across various configurations is provided in \Figref{sec:evaluation}.

\begin{figure}[t]
    \centering
    \includegraphics[width=\linewidth]{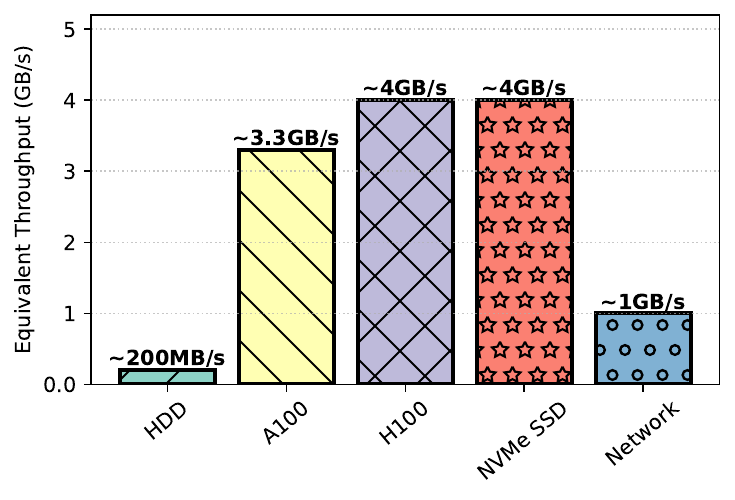}
    \caption{Comparison of equivalent \kv loading bandwidth (bytes/second) across different storage mediums and GPU computation. (Bandwidth for GPU computation is calculated by dividing the total \kv size by processing time.)}
    \label{fig:bandwidth-comparison}
\end{figure}

Our results in \Figref{fig:bandwidth-comparison} demonstrate that computation and I/O resources achieve comparable throughput in typical inference scenarios. Specifically, computing \kv with an H100 GPU achieves similar throughput to loading from SSD and is much faster than network bandwidth used in Google Cloud Egress (\verb|~|1GB/s)~\cite{googlecloudbandwidth}. This observation validates our design principle of leveraging both compute and I/O resources in parallel to minimize latency.

\textbf{Part 2: Scheduling Based on the Distinct Characteristics of Compute and I/O.}  
We design the \sys core scheduler as a bidirectional \kv loader based on the following key insight:  

\fbox{
    \parbox{0.9\linewidth}{
        \textbf{Insight:} Compute cost increases for later tokens, while I/O cost remains constant regardless of token position.
    }
}  

This is because, in the attention operation, later tokens must attend to all previous tokens, resulting in a higher computational cost as the sequence progresses. However, the size of key-value vectors remain the same regardless of position, meaning I/O access cost is uniform across all tokens.  

\begin{figure}[t]
    \centering
    \includegraphics[width=0.49\textwidth]{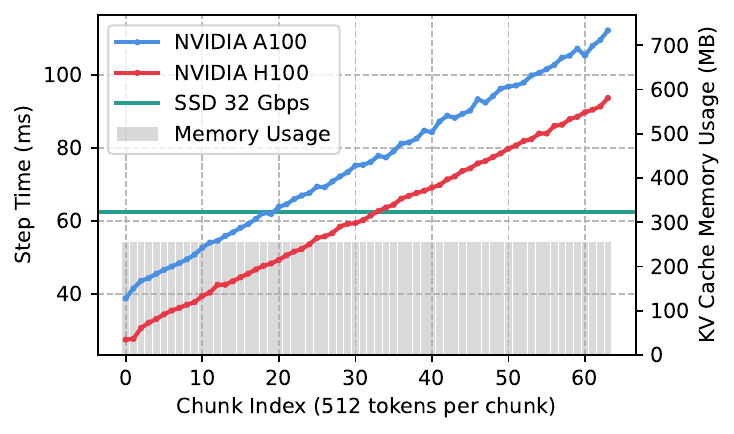}
    \caption{Chunk prefill time per step using different methods v.s. chunk index.}
    \label{fig:chunk-prefill-latency}
\end{figure}

To validate this hypothesis, we conducted long-context prefill experiments using chunked prefill, where each chunk contains a fixed 512 tokens. As shown in \Figref{fig:chunk-prefill-latency}, the results exhibit a clear pattern: latency per chunk increases linearly with its index, with earlier chunks requiring less computation, while \kv memory usage remains constant across all chunks.  

Leveraging this insight, \sys schedules earlier chunks for computation and later chunks for I/O loading. As illustrated in Figure~\ref{fig:cake-diagram}, upon receiving a request, \sys splits the sequence into chunks and initiates two parallel processes:  
(1) The local GPU computes \kv from the beginning chunk, progressing forward.  
(2) The data loading process fetches \kv from the last chunk, moving in reverse direction.  
This bidirectional strategy continues until the two processes meet in the middle, completing \kv generation for the entire prompt.

\textbf{Part 3: Practical Considerations for Real-World Deployment.}  
In practical deployment, \sys is designed to address two key challenges:  
(1) Efficiently sharing compute resources with non-prefix caching requests.  
(2) Handling fluctuations in compute and I/O availability.  

For the first challenge, real-world inference scenarios include requests without saved prefix cache that require compute resources for prefill and decoding. To ensure system throughput, \sys can prioritize other users' compute needs, allocating only available resources to \sys computation. Since \sys can still rely on I/O loading even when compute resources are constrained, we introduce an \textbf{adaptive scheduling mode} that extends vLLM’s scheduling logic as discussed in \Secref{sec:background}. This mode follows a prioritized allocation order:  
1. Decoding requests.  
2. Non-prefix-cache chunked prefill.  
3. Prefix-cache chunked prefill.  

For the second challenge, since compute resources are shared across multiple users, and I/O bandwidth could fluctuate over time, \sys must dynamically adapt to these variations while maintaining optimal latency. However, this challenge is naturally mitigated by \sys’s bidirectional parallel design—whenever I/O or compute availability changes, the merging point of the two processes shifts accordingly, always ensuring minimal latency.

\mysection{Evaluation}
\label{sec:evaluation}

In this section, we evaluate \sys across a variety of real-world setups and discuss its benefits and limitations in deployment. These insights can guide future implementations and optimizations. We begin by outlining the experiment setup and then conduct comprehensive experiments to assess the impact of the following factors on \sys:  

1. Varying I/O bandwidth and compute configurations (\Secref{sec:evaluate-compute-io}).  
2. Sequence length (\Secref{sec:sequence-len}).  
3. Model architecture (\Secref{sec:model-architecture}).  
4. \kv compression techniques (\Secref{sec:kv-cache-compression}).  
5. Resource fluctuation scenarios (\Secref{sec:fluctuation}).  
6. Adaptive Scheduling (\Secref{sec:adaptive-scheduling}).  
7. System overhead (\Secref{sec:eval-overhead}).

\subsection{Experiment Setup} \label{sec:setup}

\textbf{Models.} We evaluate \sys on various long-context models with different architectures and sizes, including LongAlpaca-7B and LongAlpaca-13B~\cite{chen2023longlora}, which are based on LLaMA 2, as well as LLaMA 3.1-8B and LLaMA 3.1-70B. Due to hardware constraints, we use the FP8-weight version for LLaMA 3.1-70B, while all other models use FP16 weights.  The first two are multi-head attention (MHA) models, while the last two apply group query attention (GQA), which will introduce different computation vs. memory. We use BF16 as the default kv cache data type. The details of the model are listed in Table~\ref{tab:model_comparison}.

\begin{table}[t]
    \centering
    \setlength{\tabcolsep}{3pt} %
    \renewcommand{\arraystretch}{0.95} %
    \small %
    \begin{tabular}{lcccc}
        \toprule
        Model & \#Layers & KV-size/Token & \#KV Hd & \#Attn Hd \\ 
        \midrule
        LongAlpaca-7B  & 32  & 512 kB & 32 & 32 \\
        LongAlpaca-13B & 40  & 800 kB & 32 & 32 \\
        LLaMA 3.1-8B  & 32 & 128 kB & 8 & 32 \\
        LLaMA 3.1-70B & 80 & 320 kB & 8 & 64 \\
        \bottomrule
    \end{tabular}
    \caption{Comparison of experimental model configs.}
    \label{tab:model_comparison}
\end{table}

\textbf{Evaluation Metrics.} We use time-to-first-token (TTFT) as our primary evaluation metric. TTFT is widely used in LLM inference, measuring the time between the arrival of a user query and the generation of the first token. In other words, it reflects either the time of loading stored \kv or computing new \kv.

\textbf{Datasets.} We evaluate \sys across various context lengths using three datasets with different task types: LongChat~\cite{li2023long} for multi-turn conversations, and TriviaQA and NarrativeQA~\cite{bai2023longbench} for long-document question-answering tasks.  According to statistical analysis from CacheGen~\cite{liu2023cachegen}, most dataset queries range between 4k and 16k tokens in length. Since specific token values do not impact \sys's performance evaluation (only token length matters), we generate synthetic prompts by uniformly sampling token lengths every 2k tokens within this range.  Additionally, as discussed, \sys is designed to optimize the latency of loading \textbf{cached} \kv from high-capacity, low-bandwidth storage layers. Thus, throughout our evaluation, we precompute and store all requests' \kv in advance.

\textbf{Baselines.} We compare \sys to two types of \kv prefill/loading mechanisms:

\BULLET Compute-only methods, which employ chunk prefill to compute all \kv. We use vLLM (v0.6.2) in chunk prefill mode with token budget sizes of 512 and 1024. 

\BULLET I/O Load-only, which loads saved \kv from local/remote disks through disk/network. We use LMCache (v0.1.4). %

\textbf{Hardware setting.} We run our evaluation on two server configurations:
1) A server equipped with two NVIDIA A100 80GB GPUs connected via NVLink, a 64-core AMD EPYC 7763 CPU, and 2.0TB of memory.  
2) A server with a single NVIDIA H100 GPU, a 26-core vCPU, and 200GB of memory. 

\textbf{I/O Bandwidth Control.}
To accurately manage I/O bandwidth and ensure reproducibility, We simulate the chunk I/O loading process by calculating the appropriate delay time based on the chunk size and network bandwidth. The simulated storage backend is then set to pause data transfer to GPU memory until the specified delay has elapsed. We choose a variety of bandwidth configurations as demonstrated in Table~\ref{tab:bw_contral}. We choose the I/O loading part chunk size as 128 tokens as it empirically strikes an optimal balance between bandwidth utilization and processing granularity.

\begin{table}[t]
    \centering
    \setlength{\tabcolsep}{3pt} %
    \renewcommand{\arraystretch}{0.95} %
    \small %
    \begin{tabular}{l|r}
        \toprule
        Physical Config & Bandwidth \\ 
        \midrule
        Google Cloud standard egress limit & 7 Gbps  \\
        Google Cloud tier-1 egress limit & 25 Gbps  \\
        Lambda Lab SSD read & 32 Gbps \\
        Samsung 980 pro SSD read & 56 Gbps \\
        Inifiniband (RoCE) & 100 Gbps \\
        \bottomrule
    \end{tabular}
    \caption{I/O bandwidth with corresponding physical configuration}
    \label{tab:bw_contral}
\end{table}

\textbf{GPU Resource Utilization.} In online serving scenarios, a single machine often handles multiple user requests concurrently. As a result, a user’s prefill operation may not always have access to the full GPU resources. To evaluate different GPU resource availability conditions, we adopt vLLM’s token budget scheduling policy, as discussed in \S\ref{sec:background}, to represent resource utilization. For instance, if the total token budget is 512 tokens and a \sys request consumes 256 tokens while the remaining tokens are allocated to other users, we define this scenario as 50\% GPU utilization. We use this definition to simulate varying levels of GPU resource utilization.

\textbf{\kv Compression methods.} \kv Compression methods, which are orthogonal to our work. They can compress the \kv size to make them more efficiently transferable through I/O. In our evaluation, we combine the most common 8-bit quantization and 3-bit quantization proposed by KVQuant~\cite{hooper2024kvquant} with \sys to further evaluate its performance.

\subsection{Evaluation Across Compute and I/O  Configurations}
\label{sec:evaluate-compute-io}
In this section, we evaluate the performance of \sys under varying computational resources and I/O settings. Table~\ref{tab:compute-vs-IO} presents the speedup achieved by \sys compared to an I/O-only approach and a compute-only approach across different GPU hardware, GPU utilization levels, and I/O configurations. We conduct tensor parallel inference using a 2×A100 setup.  

We observe that, for a fixed GPU utilization level, the speedup over I/O-only increases progressively from A100 to H100 and further to 2×A100. For example, under 100\% GPU utilization with an I/O load bandwidth of 32 Gbps (representing a common server SSD read bandwidth), the speedup over I/O-only improves from 2× on A100 to 2.23× on H100, and 2.63× on 2×A100. This aligns with our design expectation that as compute resources increase, the performance advantage over an I/O-only approach continues to grow. Similarly, analyzing different GPU utilization levels within the same hardware setup yields the same trend.  Additionally, examining Table~\ref{tab:compute-vs-IO} row by row, we observe that as I/O bandwidth increases, the speedup over compute-only also improves. This highlights that \sys effectively balances both compute and I/O resources to enhance overall performance.  

From a broader perspective, the most favorable deployment scenario for \sys occurs when its speedup is approximately 2× compared to both compute-only and I/O-only approaches. This scenario arises when computation and I/O capabilities are well-balanced, enabling \sys to efficiently utilize both in parallel and achieve the best average speedup over either baseline. We observe that in most cases, \sys provides a significant improvement, demonstrating its strong performance.

To aid interpretation, we highlight scenarios with at least a 1.5× improvement over both baselines in \textcolor{teal}{teal}, indicating cases where \sys effectively leverages both resources. Conversely, we use text in \textcolor{red}{red} to mark cases where \sys achieves more than a 10× improvement over one baseline, which suggests a severe bottleneck in the other resource. In such extreme cases, comparing with the more capable resources, \sys's speedup is limited. 

To ensure a fair evaluation of \sys’s average performance improvement over the two baseline methods, we exclude the data points highlighted in red, as comparing \sys’s speedup relative to an extremely weak baseline would be misleading. After filtering these outliers, we compute the average speedup to provide a more balanced assessment of \sys’s overall efficiency. On average, \sys achieves 2.23x and 3.76x speedup over I/O-only and compute-only methods respectively, under different bandwidth and computation utilization levels.

\begin{table}[t]
    \centering
    \setlength{\tabcolsep}{4pt} %
    \renewcommand{\arraystretch}{1.1} %
    \tiny %
    \begin{tabular}{cc|ccccc}
        \toprule
        \textbf{Hardware} & \textbf{Util} & \textbf{7 Gbps} & \textbf{25 Gbps} & \textbf{32 Gbps} & \textbf{56 Gbps} & \textbf{100 Gbps} \\
        \midrule
2xA100 & 12.5\% & \cellcolor{teal!10} 1.87\textbackslash{}2.18 & 1.25\textbackslash{}5.06 & 1.18\textbackslash{}6.12 & 1.09\textbackslash{}9.54 & \cellcolor{pink} 1.04\textbackslash{}15.27 \\
2xA100 & 50\% & 4.57\textbackslash{}1.29 & \cellcolor{teal!10} 2.07\textbackslash{}2.03 & \cellcolor{teal!10} 1.80\textbackslash{}2.24 & 1.46\textbackslash{}3.06 & 1.30\textbackslash{}4.62 \\
2xA100 & 87.5\% & 7.22\textbackslash{}1.15 & \cellcolor{teal!10} 2.75\textbackslash{}1.53 & \cellcolor{teal!10} 2.36\textbackslash{}1.67 & \cellcolor{teal!10} 1.83\textbackslash{}2.17 & \cellcolor{teal!10} 1.50\textbackslash{}3.02 \\
2xA100 & 100\% & 8.10\textbackslash{}1.10 & 3.03\textbackslash{}1.43 & \cellcolor{teal!10} 2.63\textbackslash{}1.59 & \cellcolor{teal!10} 1.97\textbackslash{}2.00 & \cellcolor{teal!10} 1.54\textbackslash{}2.64 \\
A100 & 12.5\% & \cellcolor{teal!10} 1.57\textbackslash{}2.85 & 1.14\textbackslash{}7.22 & 1.10\textbackslash{}8.95 & 1.03\textbackslash{}14.06 & \cellcolor{pink} 1.02\textbackslash{}23.59 \\
A100 & 50\% & 3.35\textbackslash{}1.46 & \cellcolor{teal!10} 1.68\textbackslash{}2.55 & \cellcolor{teal!10} 1.55\textbackslash{}3.01 & 1.30\textbackslash{}4.26 & 1.17\textbackslash{}6.48 \\
A100 & 87.5\% & 5.05\textbackslash{}1.26 & \cellcolor{teal!10} 2.18\textbackslash{}1.88 & \cellcolor{teal!10} 1.90\textbackslash{}2.10 & \cellcolor{teal!10} 1.56\textbackslash{}2.91 & 1.30\textbackslash{}4.07 \\
A100 & 100\% & 5.62\textbackslash{}1.21 & \cellcolor{teal!10} 2.41\textbackslash{}1.81 & \cellcolor{teal!10} 2.00\textbackslash{}1.91 & \cellcolor{teal!10} 1.69\textbackslash{}2.71 & 1.30\textbackslash{}3.52 \\
H100 & 12.5\% & \cellcolor{teal!10} 1.74\textbackslash{}2.40 & 1.21\textbackslash{}5.80 & 1.13\textbackslash{}6.90 & \cellcolor{pink} 1.08\textbackslash{}11.08 & \cellcolor{pink} 1.05\textbackslash{}18.24 \\
H100 & 50\% & 4.13\textbackslash{}1.38 & \cellcolor{teal!10} 1.94\textbackslash{}2.25 & \cellcolor{teal!10} 1.72\textbackslash{}2.55 & 1.45\textbackslash{}3.62 & 1.27\textbackslash{}5.34 \\
H100 & 87.5\% & 6.23\textbackslash{}1.17 & \cellcolor{teal!10} 2.53\textbackslash{}1.66 & \cellcolor{teal!10} 2.23\textbackslash{}1.86 & \cellcolor{teal!10} 1.74\textbackslash{}2.44 & \cellcolor{teal!10} 1.50\textbackslash{}3.54 \\
H100 & 100\% & 7.13\textbackslash{}1.17 & \cellcolor{teal!10} 2.74\textbackslash{}1.56 & \cellcolor{teal!10} 2.40\textbackslash{}1.75 & \cellcolor{teal!10} 1.84\textbackslash{}2.25 & 1.49\textbackslash{}3.08
 \\
        \bottomrule
    \end{tabular}
        \caption{Speedup over I/O-only \textbackslash{} compute-only methods across different compute and I/O configurations. Chunk size: 512, Model: LongAlpaca-13B, Sequence length: 16K.}
    \label{tab:compute-vs-IO}
\end{table}

\mysubsection{Evaluation Across Sequence Lengths}
\label{sec:sequence-len}
In this section, we evaluate the performance of \sys under varying sequence length settings. Table~\ref{tab:seq-len} presents the speedup achieved by \sys compared to an I/O-only approach and a compute-only approach across different compute-I/O configurations and sequence length settings.  

We observe that, for a fixed computation and I/O configuration, increasing the sequence length consistently improves the speedup over compute-only methods. As discussed in \Secref{sec:design}, later tokens in a sequence require more computation. Therefore, as sequence length grows, leveraging I/O-only methods to load tokens from later positions becomes increasingly beneficial compared to compute-only approaches, leading to a continuous increase in speedup.  

Following the interpretation aid strategy in \Secref{sec:evaluate-compute-io}, We highlight less effective scenarios in \textcolor{red}{red} and highly beneficial scenarios in \textcolor{teal}{teal}, observing that \sys is highly beneficial in many cases. Notably, there is a single scenario where \sys underperforms compared to compute-only methods. This occurs when the sequence is short, as using 2×A100 for computation is significantly faster than loading data via a 7 Gbps bandwidth. However, due to \sys’s scheduling algorithm, it still assigns a portion of the workload to the I/O-only method, introducing overhead. For future work, we can optimize this scheduling strategy by incorporating an estimation mechanism. If one resource significantly outperforms the other, \sys could adaptively fall back to a single-resource mode, utilizing only the more efficient resource to minimize overhead.  

However, on average, we can still observe \sys improves prefilling speed 3.34x faster than the I/O-only method and 2.24x faster than the compute-only method.

\begin{table}[t]
    \centering
    \setlength{\tabcolsep}{4pt} %
    \renewcommand{\arraystretch}{1.2} %
    \tiny %
    \begin{tabular}{ccc|cccc}
        \toprule
\textbf{Hardware} & \textbf{BW} & \textbf{Util} & \textbf{4k Tokens} & \textbf{8k Tokens} & \textbf{12k Tokens} & \textbf{16k Tokens} \\
\midrule
2xA100 & 7Gbps & 12.5\% & \cellcolor{teal!10} 2.02\textbackslash{}1.83 & \cellcolor{teal!10} 1.97\textbackslash{}1.95 & \cellcolor{teal!10} 1.94\textbackslash{}2.09 & \cellcolor{teal!10} 1.87\textbackslash{}2.18 \\
2xA100 & 7Gbps & 50\% & 4.75\textbackslash{}1.08 & 4.97\textbackslash{}1.21 & 4.71\textbackslash{}1.24 & 4.57\textbackslash{}1.29 \\
2xA100 & 7Gbps & 87.5\% & 7.31\textbackslash{}1.00 & 7.49\textbackslash{}1.06 & 7.61\textbackslash{}1.15 & 7.22\textbackslash{}1.15 \\
2xA100 & 7Gbps & 100\% & \cellcolor{pink} 9.06\textbackslash{}0.99 & 8.65\textbackslash{}1.04 & 8.43\textbackslash{}1.08 & 8.10\textbackslash{}1.10 \\
2xA100 & 32Gbps & 12.5\% & 1.13\textbackslash{}4.41 & 1.19\textbackslash{}5.17 & 1.19\textbackslash{}5.68 & 1.18\textbackslash{}6.12 \\
2xA100 & 32Gbps & 50\% & \cellcolor{teal!10} 1.68\textbackslash{}1.64 & \cellcolor{teal!10} 1.87\textbackslash{}1.99 & \cellcolor{teal!10} 1.80\textbackslash{}2.09 & \cellcolor{teal!10} 1.80\textbackslash{}2.24 \\
2xA100 & 32Gbps & 87.5\% & 2.47\textbackslash{}1.45 & \cellcolor{teal!10} 2.45\textbackslash{}1.52 & \cellcolor{teal!10} 2.49\textbackslash{}1.66 & \cellcolor{teal!10} 2.36\textbackslash{}1.67 \\
2xA100 & 32Gbps & 100\% & 2.67\textbackslash{}1.25 & 2.73\textbackslash{}1.43 & \cellcolor{teal!10} 2.69\textbackslash{}1.52 & \cellcolor{teal!10} 2.63\textbackslash{}1.59 \\
1xA100 & 7Gbps & 12.5\% & \cellcolor{teal!10} 1.62\textbackslash{}2.30 & \cellcolor{teal!10} 1.59\textbackslash{}2.47 & \cellcolor{teal!10} 1.58\textbackslash{}2.67 & \cellcolor{teal!10} 1.57\textbackslash{}2.85 \\
1xA100 & 7Gbps & 50\% & 3.58\textbackslash{}1.27 & 3.48\textbackslash{}1.32 & 3.45\textbackslash{}1.41 & 3.35\textbackslash{}1.46 \\
1xA100 & 7Gbps & 87.5\% & 5.36\textbackslash{}1.12 & 5.18\textbackslash{}1.15 & 5.17\textbackslash{}1.21 & 5.05\textbackslash{}1.26 \\
1xA100 & 7Gbps & 100\% & 6.31\textbackslash{}1.11 & 5.95\textbackslash{}1.13 & 5.66\textbackslash{}1.15 & 5.62\textbackslash{}1.21 \\
1xA100 & 32Gbps & 12.5\% & 1.04\textbackslash{}6.37 & 1.06\textbackslash{}7.20 & 1.08\textbackslash{}8.04 & 1.10\textbackslash{}8.95 \\
1xA100 & 32Gbps & 50\% & 1.38\textbackslash{}2.10 & 1.49\textbackslash{}2.48 & \cellcolor{teal!10} 1.50\textbackslash{}2.71 & \cellcolor{teal!10} 1.55\textbackslash{}3.01 \\
1xA100 & 32Gbps & 87.5\% & \cellcolor{teal!10} 1.83\textbackslash{}1.65 & \cellcolor{teal!10} 1.93\textbackslash{}1.87 & \cellcolor{teal!10} 1.91\textbackslash{}1.97 & \cellcolor{teal!10} 1.90\textbackslash{}2.10 \\
1xA100 & 32Gbps & 100\% & \cellcolor{teal!10} 2.06\textbackslash{}1.55 & \cellcolor{teal!10} 2.15\textbackslash{}1.79 & \cellcolor{teal!10} 2.00\textbackslash{}1.79 & \cellcolor{teal!10} 2.00\textbackslash{}1.91 \\
        \bottomrule
    \end{tabular}
        \caption{Speedup over I/O-only \textbackslash{} compute-only methods across different sequence length settings, Hardware: 2xA100, Chunk size 512, Model: Long-Alpaca-13B}
    \label{tab:seq-len}
\end{table}
\vspace{-10pt}

\mysubsection{Evaluation Across Model Architectures}
\label{sec:model-architecture}

Table~\ref{tab:model-architecture} presents the speedup achieved by \sys compared to an I/O-only approach and a compute-only approach across different compute-I/O configurations and model architectures. Both LongAlpaca models employ the Multi-Head Attention (MHA) mechanism, while the LLaMA 3 series utilizes Grouped Query Attention (GQA) to compress the KV cache, thereby reducing \kv memory overhead.  

By comparing the LongAlpaca-7B and LLaMA 3.1-8B models under the same compute and I/O settings, we observe that, despite the I/O bandwidth remaining unchanged, the speedup relative to the compute-only method increases. This is due to the benefits of GQA, which reduces the \kv size, effectively enhancing I/O efficiency.  

Furthermore, comparing different model sizes within the same series reveals that the speedup for I/O-only methods decreases as the model size grows. This indicates that computation requirements scale faster than I/O demands, leading to diminishing relative gains from compute optimizations.  

We also identified four data points where \sys did not outperform the baseline, similar to the issue discussed in \Secref{sec:sequence-len}. This can be mitigated by incorporating a fallback mechanism that dynamically adjusts resource allocation when one resource significantly outperforms the other. Despite the overheads on extreme scenarios, \sys on average achieves 2.68x speedup over the I/O-only method and 3.01 speedup over the compute-only method. 

\begin{table}[h!]
    \centering
    \setlength{\tabcolsep}{4pt} %
    \renewcommand{\arraystretch}{1.1} %
    \tiny %
    \begin{tabular}{cc|ccccc}
        \toprule
\textbf{Util} & \textbf{BW} & \textbf{LongAlpaca-7B} & \textbf{LongAlpaca-13B} & \textbf{Llama3.1-8B} & \textbf{Llama3.1-70B} \\
\midrule
12.5\% & 7Gbps & \cellcolor{teal!10} 1.92\textbackslash{}2.15 & \cellcolor{teal!10} 1.87\textbackslash{}2.18 & 1.20\textbackslash{}5.56 & \cellcolor{pink} 0.99\textbackslash{}12.92 \\
12.5\% & 32Gbps & 1.20\textbackslash{}5.98 & 1.18\textbackslash{}6.12 & \cellcolor{pink} 0.98\textbackslash{}19.47 & \cellcolor{pink} 0.80\textbackslash{}44.26 \\
50\% & 7Gbps & 4.79\textbackslash{}1.30 & 4.57\textbackslash{}1.29 & \cellcolor{teal!10} 1.88\textbackslash{}2.13 & 1.25\textbackslash{}3.99 \\
50\% & 32Gbps & \cellcolor{teal!10} 1.85\textbackslash{}2.22 & \cellcolor{teal!10} 1.80\textbackslash{}2.24 & 1.13\textbackslash{}5.42 & \cellcolor{pink} 0.91\textbackslash{}12.24 \\
100\% & 7Gbps & 8.55\textbackslash{}1.10 & 8.10\textbackslash{}1.10 & \cellcolor{teal!10} 2.81\textbackslash{}1.52 & \cellcolor{teal!10} 1.66\textbackslash{}2.64 \\
100\% & 32Gbps & \cellcolor{teal!10} 2.79\textbackslash{}1.60 & \cellcolor{teal!10} 2.63\textbackslash{}1.59 & 1.37\textbackslash{}3.18 & 1.02\textbackslash{}6.86 \\
        \bottomrule
    \end{tabular}
    
        \caption{Speedup over I/O-only \textbackslash{} compute-only across different model settings, Hardware: 2xA100, Chunk size 512, Seq-len: 16k.}
    \label{tab:model-architecture}
\end{table}
\vspace{-10pt}

\begin{table}[h!]

    \centering
    \setlength{\tabcolsep}{8pt} %
    \renewcommand{\arraystretch}{1.2} %
    \tiny %
    \begin{tabular}{c|ccc}
        \toprule
\textbf{BW} & \textbf{16-bit} & \textbf{8-bit} & \textbf{3-bit} \\
\midrule
7Gbps & 8.10\textbackslash{}1.10 & 4.63\textbackslash{}1.22 & 2.39\textbackslash{}1.63 \\
25Gbps & 3.03\textbackslash{}1.43 & 2.07\textbackslash{}1.87 & 1.41\textbackslash{}3.09 \\
32Gbps & 2.63\textbackslash{}1.59 & 1.85\textbackslash{}2.12 & 1.37\textbackslash{}3.73 \\
56Gbps & 1.97\textbackslash{}2.00 & 1.49\textbackslash{}2.85 & 1.26\textbackslash{}5.38 \\
        \bottomrule
    \end{tabular}
        \caption{Speedup over I/O-only \textbackslash{} compute-only under different low-precision compression, Hardware: 2xA100 100\% Utilization, Chunk size 512, Seq-len: 16k, Model: Long-Alpaca-13B.}
    \label{tab:compression}
    \vspace{-5pt}
\end{table}

\subsection{Incorporating \kv Compression with \sys}
\label{sec:kv-cache-compression}

Table~\ref{tab:compression} presents the speedup achieved by \sys compared to the I/O-only approach and compute-only approach across different I/O configurations and different compression ratio. We can observe similar to the effect of GQA, low-precision compression ratio reduces the \kv size, thus enhancing the I/O load performance, thus \sys speedup over compute-only methods continues to increase as the precision decreases. 

\subsection{Handling Fluctuations in Resources}
\label{sec:fluctuation}

To evaluate \sys's performance under fluctuating compute and I/O conditions, we randomly sample a compute budget trace uniformly between 0–512 tokens to represent varying computational power. Similarly, we randomly sample an I/O bandwidth trace between 0–25 Gbps to assess how \sys adapts its scheduling strategy.  

\begin{figure}[t]
    \centering
    \includegraphics[width=\linewidth]
    {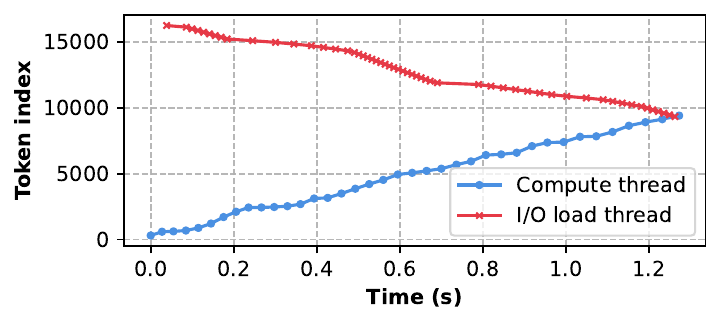}
    \caption{\sys trace under fluctuate network and available computation power. Hardware: A100, Model: Long-Alpaca-7B, I/O Bandwidth: 0-25Gbps, Compute Utilization: 0-512 budget, Seq-len: 16k.}
    \label{fig:dynamic-network}
\end{figure}

As shown in \Figref{fig:dynamic-network}, \sys dynamically leverages both I/O and computational resources, regardless of fluctuations. Its bidirectional prefetching mechanism automatically identifies the optimal merging point to minimize TTFT, ensuring optimal performance even under varying resource constraints.

\subsection{Evaluation on Adaptive Scheduling}
\label{sec:adaptive-scheduling}

In this evaluation, we give an example to evaluate the performance of the adaptive scheduling algorithm in \sys. We begin by sending a prefix-caching request of length 16K, representing a typical long-context prefix-caching scenario in \sys. To simulate a burst of incoming requests from other users, we then randomly generate 22 additional requests with sequence lengths ranging from 32 to 448, following a spiked distribution.  

\begin{figure}[t]
    \centering
    \includegraphics[width=\linewidth]
    {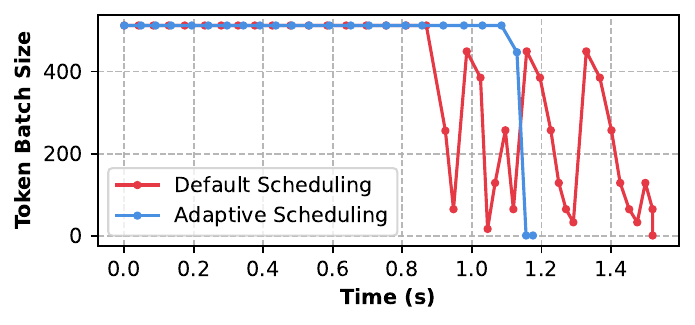}
    \caption{Token batch size over time under dynamic workload with default and adaptive scheduling. Adaptive scheduling of \sys flattens the spikes of dynamic workloads thus maximizes the system throughput.}
    \label{fig:scheduling}
\end{figure}

Using these request traces, we compare the default vLLM scheduling algorithm with \sys's adaptive scheduling algorithm for inference execution. The result is demonstrated in \Figref{fig:scheduling}. In vLLM's default scheduling mode, our prefix-caching request arrives first and is processed to completion before handling subsequent non-prefix-caching requests. This approach results in suboptimal GPU utilization, as the token batch size is not fully utilized at all times.  

In contrast, \sys’s adaptive scheduling prioritizes decoding and prefill operations for incoming non-prefix-caching requests while allocating the remaining compute budget for chunk prefill of the prefix-caching request. It successfully keeps the GPU busy and reduces the overall finish time from 1.5s to 1.19s, improving throughput by 26\%.  

\subsection{Overheads of \sys }
\label{sec:overhead}

\begin{figure}[t]
    \centering
    \includegraphics[width=\linewidth]
    {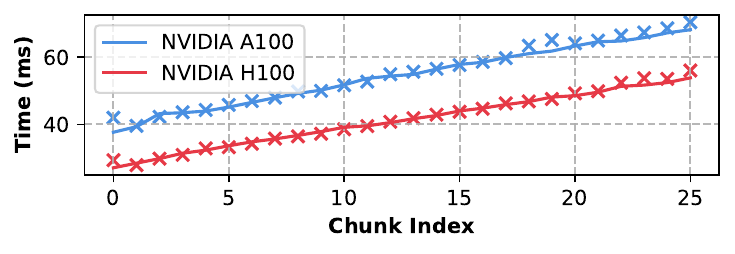}
    \caption{Per-step inference time in vLLM before and after integration with \sys. The solid line represents the step time without \sys, while the 'x' markers indicate step times with \sys.}
    \label{fig:fig7a}
    \vspace{-0.1in}
\end{figure}

\label{sec:eval-overhead}

To evaluate the overhead introduced by \sys, we compare the duration of each engine step between the original vLLM and vLLM with \sys. As shown in \Figref{fig:fig7a}, we launch a chunk prefill job on both A100 and H100 servers. The chunk prefill time of vLLM with \sys closely follows the trace of the original vLLM, indicating minimal performance impact.  

These results demonstrate that \sys introduces negligible overhead, as it only performs a lightweight check to determine whether the next chunk has been fetched at runtime.

\mysection{Conclusion}

In this paper, we introduced \sys, the first KV cache loading system that optimally balances computation and I/O to minimize Time to First Token (TTFT) in LLM inference. Unlike prior approaches that focus on either compute or I/O in isolation, \sys employs a bidirectional scheduling strategy that dynamically adapts to resource availability, achieving an average \textbf{2.6$\times$} TTFT reduction compared to baselines. Additionally, its adaptive scheduling mechanism enhances system throughput, making it a practical and easily deployable solution for LLM-serving systems. Through extensive evaluations, we provide a detailed analysis of the scenarios where \sys is most beneficial, offering valuable insights for real-world deployment.

\newpage

\bibliography{reference}
\bibliographystyle{icml2025}

\newpage
\appendix
\onecolumn
\mysection{Details of \sys Algorithm}
\label{sec:algorithm}

The workflow of \sys can be described in detail as follows:

1. Upon receiving a request, \sys first splits the input sequence into chunks of a predetermined size.

2. Compute the prefix hash of all the chunks and find the latest prefix hash that exists in the storage backend, and determine the $total\_tokens$.

2. Two pointers are initialized: $compute\_ptr$ starting at the beginning of the sequence (index 0), and $io\_ptr$ at the end of the sequence (index $total\_tokens - 1$).

3. Two parallel processes are initiated:
   a) A GPU computation thread that starts from $compute\_ptr$ and moves forward.
   b) An I/O streaming thread that starts from $io\_ptr$ and moves backward.

4. The GPU computation thread:
   - Computes \kv for chunks starting from $compute\_ptr$.
   - After each chunk computation, it updates $compute\_ptr$ by adding the chunk size.
   - Continues until $compute\_ptr$ reaches or surpasses $io\_ptr$, or until the required \kv is found in CPU memory.

5. The I/O streaming thread:
   - Fetches pre-computed \kv for chunks ending at $io\_ptr$ from storage (local or remote) to CPU memory.
   - After each chunk fetch, it updates $io\_ptr$ by subtracting the chunk size.
   - Continues until $io\_ptr$ reaches or goes below $compute\_ptr$.

6. The process concludes when the two pointers meet or cross each other, indicating that \kv for the entire sequence has been either computed or loaded.

7. Finally, \sys returns the complete \kv for the entire sequence, ready for use in the subsequent inference steps.

This bidirectional approach allows \sys to efficiently utilize both computational and I/O resources simultaneously, minimizing idle time and optimizing the overall latency of \kv preparation for long-context LLM inference.

\begin{algorithm}
    \caption{\sys Bidirectional \kv Loading Algorithm}
    \begin{algorithmic}[1]
    \small
    \Procedure{ComputeKV}{}
        \While{$compute\_ptr < io\_ptr$}
            \If{\Call{IsInCPUMemory}{$compute\_ptr$, $COMP\_CHUNK\_SIZE$}}
                \State Signal I/O worker to stop
                \State \textbf{break}
            \EndIf
            \State Compute \kv for chunk starting at $compute\_ptr$ using GPU
            \State $compute\_ptr \gets compute\_ptr + COMP\_CHUNK\_SIZE$
        \EndWhile
    \EndProcedure
    
    \Procedure{FetchKV}{}
        \While{$compute\_ptr < io\_ptr$}
            \State Fetch \kv for chunk ending at $io\_ptr$ from storage to CPU Memory
            \State $io\_ptr \gets io\_ptr - FETCH\_CHUNK\_SIZE$
        \EndWhile
    \EndProcedure
    
    \State Initialize CPU Memory, $compute\_ptr = 0$, $io\_ptr = total\_tokens - 1$
    \State Start \Call{ComputeKV}{} in a new thread
    \State Start \Call{FetchKV}{} in a new thread
    \State Wait for both threads to complete
    \State \Return \kv for the entire sequence
    \end{algorithmic}
    \end{algorithm}

\mysection{Implementation}
\label{sec:implementation}

We implement \sys by extending LMCache~\cite{lmcache} and integrating it with vLLM~\cite{kwon2023efficient}, adding approximately 1,000 lines of code.

\subsection{Enhancements to LMCache}
LMCache, originally developed as the \kv management backend for CacheGen~\cite{liu2023cachegen}, hashes token chunks into keys for efficient \kv retrieval. To enable \sys to continuously receive \kv in the background, we introduce the following enhancements:

\paragraph{Asynchronous Retrieval} We implement an asynchronous get operation to complement LMCache's existing asynchronous put functionality. This involves creating a dedicated worker thread that continuously reads chunk keys from a task queue and retrieves the corresponding \kv to memory. Upon successful retrieval, the chunk's key is added to a resident dictionary for quick access.

\paragraph{Buffer Preallocation} We modify LMCache to preallocate chunk buffers as soon as a chunk key is pushed to the queue. This optimization allows the worker to immediately write received \kv into memory and proceed to the next chunk without delay.

\subsection{Integration with LLM Serving Systems}
\sys operates concurrently with LLM serving systems like vLLM. The integration process works as follows:

1. Upon receiving a request, \sys divides it into chunks based on the scheduled token budget.

2. Hashed keys for these chunks are pushed to the task queue in reverse order and call \textit{batch\_retrieve} API.

3. While the asynchronous get worker fetches \kv from the end of the sequence, vLLM begins chunk prefill from the start.

4. After each vLLM engine step, \sys checks if the next chunk of tokens is already in the resident dictionary using the \textit{is\_resident} API.

5. If the chunk is resident, \sys interrupts the chunk prefill process and directs vLLM to begin token generation.

6. If the chunk is not resident, chunk prefill continues until it encounters a chunk present in the dictionary.

This bidirectional approach allows \sys to efficiently utilize both I/O and computational resources, potentially reducing the Time To First Token (TTFT) for long-context LLM inference tasks.

\end{document}